\documentclass{article}

\usepackage{arxiv}

\usepackage[utf8]{inputenc} 
\usepackage[T1]{fontenc}    
\usepackage{hyperref}       
\usepackage{url}            
\usepackage{booktabs}       
\usepackage{amsfonts}       
\usepackage{nicefrac}       
\usepackage{microtype}      
\usepackage{lipsum}
\usepackage{graphicx}
\graphicspath{ {./images/} }

\title{{FecTek{:} Enhancing Term Weight in Lexicon-Based Retrieval with Feature Context and Term-level Knowledge}}

\author{
 Zunran Wang \\
  Huawei Poisson Lab\\
  ShenZhen, China \\
  \texttt{zran.wang@outlook.com} \\
   \And
Zhonghua Li \\
  Huawei Poisson Lab\\
  ShenZhen, China \\
  \texttt{lizhonghua3@huawei.com} \\
  \And
 Wei Shen \\
  Huawei Poisson Lab\\
  ShenZhen, China \\
  \texttt{shenwei32@huawei.com} \\
  \AND
   Qi Ye \\
  Huawei Poisson Lab\\
  ShenZhen, China \\
  \texttt{ye.qi@huawei.comm} \\
  \And
   Liqiang Nie \\
  Harbin Institute of Technology\\
  ShenZhen, China \\
  \texttt{nieliqiang@gmail.com} \\
}

\begin{document}
\maketitle
\begin{abstract}
Lexicon-based retrieval has gained siginificant popularity in text retrieval due to its efficient and robust performance. To further enhance performance of lexicon-based retrieval, researchers have been diligently incorporating state-of-the-art methodologies like Neural retrieval and text-level contrastive learning approaches. Nonetheless, despite the promising outcomes, current lexicon-based retrieval methods have received limited attention in exploring the potential benefits of feature context representations and term-level knowledge guidance. In this paper, we introduce an innovative method by introducing FEature Context and TErm-level Knowledge modules(FecTek). To effectively enrich the feature context representations of term weight, the Feature Context Module (FCM) is introduced, which leverages the power of BERT's representation to determine dynamic weights for each element in the embedding. Additionally, we develop a term-level knowledge guidance module (TKGM) for effectively utilizing term-level knowledge to intelligently guide the modeling process of term weight. Evaluation of the proposed method on MS Marco benchmark demonstrates its superiority over the previous state-of-the-art approaches.
\end{abstract}


\section{Introduction}
Lexicon-based retrieval is widely used in existing IR systems due to its robustness and impressive performance. However, conventional lexicon-based retrieval methods, such as BM25, heavily depend on frequency-based term weight estimation for calculating the matching score between queries and passages. This approach frequently encounters challenges in adequately capturing context representations for term weight~\cite{zhuang2021fast}. In recent years, researchers have been actively working towards enhancing context representation for term weight without term-level labelling in Lexicon-based retrieval. Their dedication has led to significant advancements in this field.

\begin{figure}[t]
    \centering
    \includegraphics[width=0.6\textwidth]{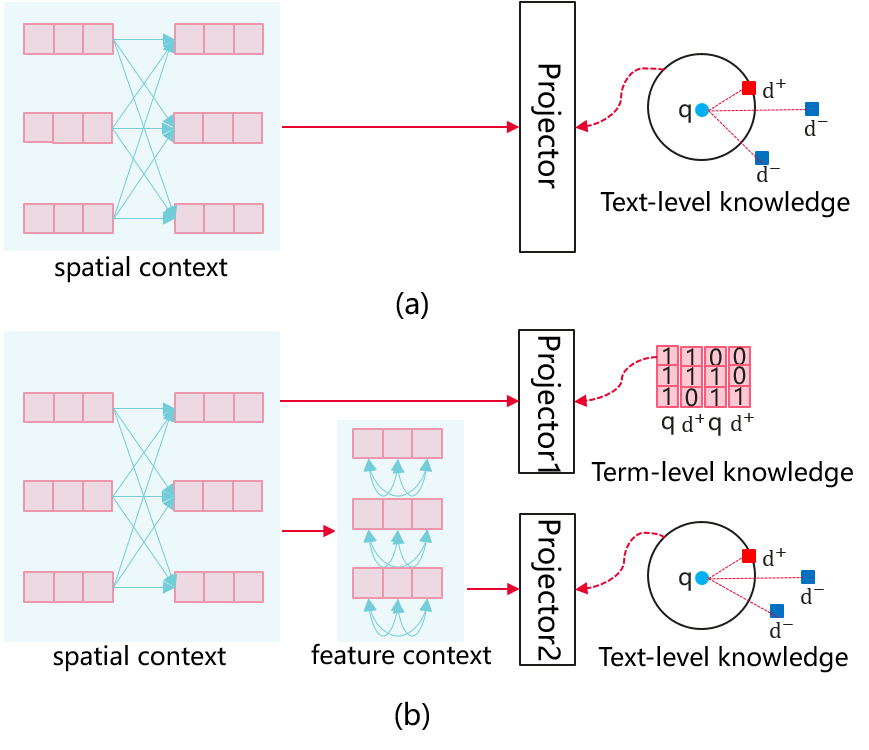}
    \caption{Lexicon-based retrieval learning architectures. (a) The mainstream architectures. (b) Our FecTek architectures: introducing feature context representations and term-level knowledge guidance. }
    \label{fig:contraction}
\end{figure}

Neural retrieval methods have emerged as a solution to tackle the issue of limited context representation, while employing a text-level contrastive learning approach to eliminate the requirement for term-level labelling. To enhance context representation, the preliminary researches leverage deep language models to produce the context representation, such as BERT~\cite{devlin2018bert}. For example, DeepCT~\cite{dai2019context} utilizes BERT to map the context term representations into term weights, and matching scores are determined by multiplying the weights of terms shared between the query and passage, followed by summing the resulting products. Numerous empirical results from DeepCT and its variants~\cite{choi2022spade,formal2021splade,mallia2021learning,zhuang2021tilde} consistently demonstrate the significant superiority of these neural retrieval methods. Nevertheless, it is worth noting that these methods primarily emphasize capturing spatial context representations, while neglecting the importance of feature context representations. To eliminate the requirement for term-level labelling, the text-level contrastive learning approach is introduced to minimize the distance between the query and the positive passage, while maximizing the distance between the query and the negative passage. This distinctive learning paradigm only requires labelling query-passage pairs, known as text-level labelling. As a result, this method has gained popularity among researchers in Lexicon-based retrieval. For instance, in the study conducted by the authors~\cite{bai2020sparterm}, an end-to-end term weight learning framework was designed by optimizing the ranking objective. Other similar approaches include UNICOIL~\cite{lin2021few}, SPLADE~\cite{choi2022spade,formal2021splade} and TILDE~\cite{zhuang2021fast,zhuang2021tilde}. While these methods effectively eliminate the need for term-level labelling, they lack clear guidance derived from term-level knowledge. Considering these findings, These observations underscore the need for improvement in lexicon-based retrieval, primarily driven by two challenges: (i) the disregarding of feature context representations, and (ii) the absence of term-level knowledge guidance.

In this paper, we present an innovative method called FecTek, as shown in Fig.~\ref{fig:contraction}, with the aim of enhancing feature context representations and incorporating term-level knowledge guidance. To address the first challenge, we devised a feature context module (FCM) inspired by the remarkable improvements achieved through the application of channel attention in CNN models~\cite{hu2018squeeze}. This module enriches the feature context representations of term weight effectively. Within the FCM, BERT's representation is utilized to determine the dynamic weights for each element in the embedding. The feature context representations are then obtained by taking the dot product between BERT's representations and these dynamic weights. The introduction of the FCM empowers FecTek to capture feature context representations.

Regarding the second problem, we developed a term-level knowledge guidance module (TKGM) as the central solution in FecTek. In lexicon-based retrieval, term weights are stored in a regular inverted index for efficient retrieval during indexing. Obviously, only term weights from the shared term between the query and passage are used to calculate the matching score. Upon observation, it becomes clear that these shared terms carry greater importance. Consequently, we integrated this observation into our approach to construct term-level knowledge. Terms found in both the query and passage are assigned a label of 1, while the remaining terms are labeled as 0. The cross-entropy loss is utilized to effectively leverage the term-level knowledge and guide the modelling process of term weight. 
 
Our main contributions of this paper are summarized as follows: 
\begin{itemize}
  \item We design an innovative approach named FecTek to elevate term weight in lexicon-based retrieval by introducing term-level knowledge and feature context.
  \item We have developed two specialized components, known as FCM and TKGM, to ensure the FecTek works effectively.
  \item Our proposed method has consistently outperformed baseline algorithms on the MS-Marco benchmark, establishing itself as the new state-of-the-art approach.
\end{itemize}

\begin{figure*}[t]
    \centering
    \includegraphics[width=1.0\textwidth]{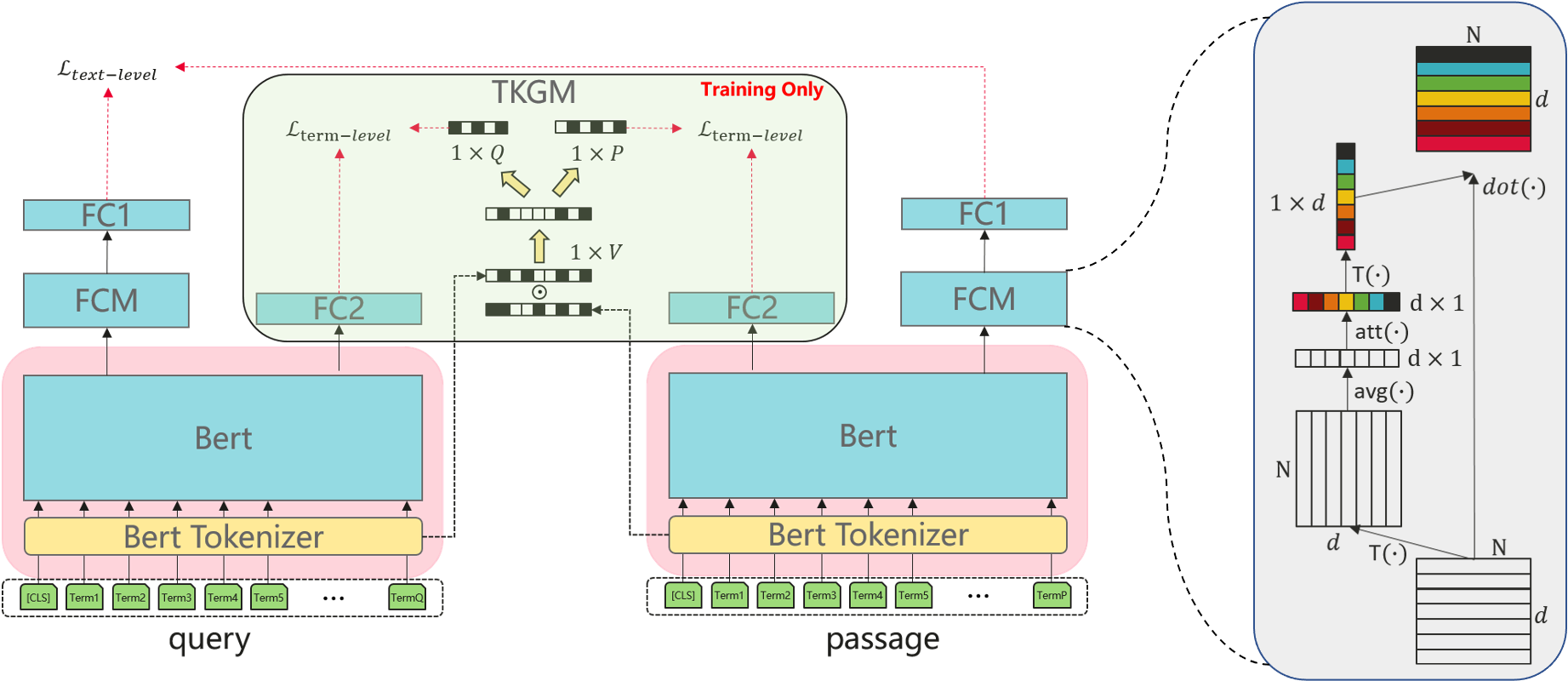}
    \caption{The architecture of our FecTek. FecTek leverages BERT to extract spatial context representations of each token. It then incorporates two branches to learn text-level and term-level knowledge respectively. The text-level branch consists of the FCM and the Projector module, while the term-level branch includes the Indicator module and an additional Projector module. Only during the training process does the term-level branch occur.}
    \label{fig:framework}
\end{figure*}

\section{Related Work}
\subsection{Dense-vector Retriever}
Dense-vector encoding methods provide a straightforward approach to represent documents/queries as dense vectors. The match score between a document and a query is then computed using either dot-product or cosine similarity. In the domain of dense vector techniques, the primary focus lies in improving representation and selecting optimal representations. To enhance the retriever's representation, mainstream methods include selecting hard training negatives and leveraging pre-training. ANCE~\cite{xiong2020approximate} proposed a learning mechanism that globally selects hard training negatives from the entire corpus, using an asynchronously updated ANN index. Conversely, ADORE~\cite{zhan2021optimizing} incorporated dynamic sampling to adapt hard negative samples during model training. Condenser~\cite{gao2021condenser} designed a pre-training strategy
tailored for ad-hoc retrieval and coCondenser~\cite{gao2021unsupervised} integrated an unsupervised corpus-level contrastive loss to refine the passage embedding space, thereby enhancing Condenser's performance. Different from methods that only rely on [CLS] vectors for computing relevance scores, ColBERT~\cite{khattab2020colbert} used fine-granular interaction to capture fine grained similarity between tokens, while COIL~\cite{gao2021coil} used word-bag match to assess the semantic lexical similarity between tokens.

\subsection{Contextual Information for Lexicon-base Retriever}
Lexicon-based retrieval has gained widespread recognition in text retrieval because of its remarkable efficiency. BM25 relies on freqency-based term weight estimation to compute the matching score between queries and passages, making it a popular choice in the industry due to its simplicity. Unfortunately, this approach often fails to adequately capture contextual information in term weight for
Lexicon-based retrieval~\cite{zhuang2021fast}. Thus, in recent years, researchers have been actively striving to improve context representation in term weight. DeepCT~\cite{dai2019context} maps the context term representations from BERT into term weights, matching scores are derived by multiplying the weights of terms shared between the query and passage, and then summing the resulting products. SparTerm~\cite{bai2020sparterm} introduced a contextual importance predictor that accurately predicts the significance of each term within the vocabulary. Moreover, a sophisticated gating controller was specifically devised to regulate the activation of terms. These two expertly crafted modules seamlessly collaborate to achieve an unmatched level of sparsity and flexibility in the text representation, effectively merging the techniques of both term-weight and expansion into a unified framework. Leveraging contextual term representations, SPLADE~\cite{formal2021splade} additionally introduced a remarkable log-saturation effect that effectively controls term dominance, ensuring natural sparsity in the resulting representations. TILDE~\cite{zhuang2021tilde} proposed a more efficient framework of lexicon-base retriever by introducing query likelihood component, thus TILDE abolished the requirement of the inference step of deep language models based retrieval approaches. However, these methods primarily emphasize capturing spatial context representations, while neglecting the importance of feature context representations.

\subsection{Term-level Label Assignment for Lexicon-base Retriever}
Text-level label assignment has gained increasing popularity among researchers in Lexicon-based retrieval due to its incredible performance. However, when it comes to the utilization of term-level label assignment in the retrieval field, we find only a handful of notable contributions. DeepCT~\cite{dai2019context} used the frequency of term as term weights, then minimized the mean square error between the predicted weight and ground truth weights. To enhance the execution of TextRank, Biased TextRank ~\cite{kazemi2020biased} introduced a captivating modification to the random restarts. This involved leveraging the vector similarity of an embedding model to regard as the relevance of two terms, i.e., the vector similarity of an embedding model was defined as the term weights. TILDE~\cite{zhuang2021tilde}, which is similar to our work, employed a labeling mechanism where terms present in both the query and passage were assigned with a label of 1, while the remaining terms were labeled as 0. However, TILDE approached the representation mapping to the vocabulary space and considered it as a multi-classification task rather than a binary classification task. Additionally, TILDE prioritized the multi-classification task as the primary objective, without incorporating a text-level contrastive learning task as the primary objective.

\subsection{Reranker-taught Retriever}
The practice of distilling scores from a reranker into a retriever has yielded promising results. 
Through examining the influence of distillation, hard-negative mining, and Pre-trained Language Model initialization, the authors of ~\cite{formal2022distillation} extended SPLADE that can reap the identical training enhancements observed in dense models. RocketQAv2~\cite{ren2021rocketqav2} was designed to establish consensus through the use of reranker-filtered hard negatives, while AR2~\cite{zhang2021adversarial}, on the contrary, adopted an adversarial learning approach, treating the retriever as a generator and the reranker as a discriminator. This enables them to optimize their performance through Reranker-taught distilling. 

\section{Methodology}
Our FecTek is tailored for lexicon-based retrieval by introducing both term-level knowledge and feature context, where term weights are calculated and then stored in a regular inverted index, for facilitating retrieval during indexing. 

\subsection{Network Architecture}
The architecture of our FecTek is illustrated in Fig.~\ref{fig:framework}. Similar to other widely adopted frameworks, the powerful BERT~\cite{devlin2018bert} serves as the backbone, generating spatial context representations($ H = [h_{cls}, h_1, h_2, ...,h_{N-2}, h_{SEP}]$). To improve the performance of lexicon-based retrieval, we have introduced two branches specifically designed to capture text-level and term-level knowledge respectively. The text-level branch, consisting of the FCM (detailed in Sec.\ref{FCM}) and projector module, is responsible for acquiring term weights. The FCM module enhances token representation through feature context learning, while the projector module skillfully transforms the token representation into the term weight space($W_{term_i}$). Conversely, the term-level branch, named TKGM(detailed in Sec.\ref{TKGM}), consists of the Indicator and projector module. The Indicator module generates term-level labels while the projector module converts the token representation into the classification space($P_{term_i}$).

\begin{figure}[t]
    \centering
    \includegraphics[width=0.2\textwidth]{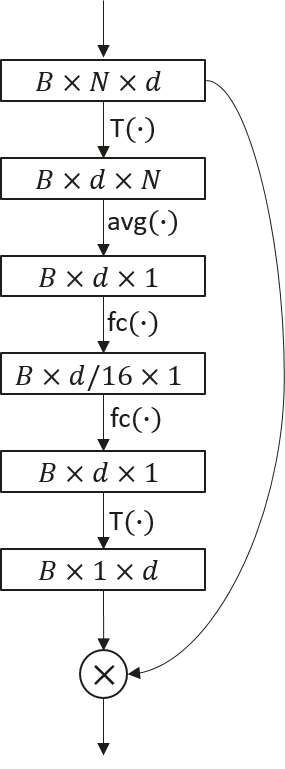}
    \caption{The feature context module. $\otimes$ denotes the point-wise multiplication.}
    \label{fig:FCM}
\end{figure}
\subsection{FCM}\label{FCM}
Following BERT with spatial context representations, the FCM module brings feature context learning. The FCM module is proposed inspired by the remarkable gains achieved through the application of channel attention in CNN models~\cite{hu2018squeeze}. As shown in Fig.~\ref{fig:FCM}, The embedding first passes by a transpose operation, which transforms their size from $B \times N \times d$ to $B \times d \times N$. The embedding is then sent into global average pooling, resulting in size changes into $B \times d \times 1$. Next, the embedding is fed into two FC layers to generate feature context. The output dimensions of them are $1/16\times$ and $1 \times$, respectively. Finally, the feature context passes by a transpose operation again, which converts their size from $B \times d \times 1$ to $B \times 1 \times d$. Consequently, the raw input embedding of the FCM is multiplied by the feature context. Embedding enhanced by feature context is then got. 

After the FCM, FecTek then combines a term's feature context representations into a term importance score~\cite{formal2021splade}:
\begin{equation}
W_{term_i} = log(1 + ReLU(w_1f_i + b_1))
\end{equation}

where $w_1$ and $b_1$ are linear weights and bias of Projector1 module, respectively. $f_i$ is $ith$ term feature context representations, i.e.,
\begin{equation}
F = FCM(H)
\end{equation}
where $FCM$ is the operation of the FCM module and $F = [f_{cls}, f_1, h_2, ..., f_{N-2}, f_{SEP}]$.

\subsection{TKGM}\label{TKGM}
The TKGM module is an auxiliary task to introduce the term-level knowledge for our FecTek, including two sub-tasks, i.e., generating supervised labels and outputs of each term.

The Indicator module is introduced to acquire supervised labels for each term. In lexicon-based retrieval, only term weights from the shared term between the query and passage are used to calculate the matching score. Upon observation, it becomes clear that these shared terms carry greater importance. Consequently, we integrated this observation into our approach to introduce the Indicator module. Terms present in both the query and passage are designated with a label of 1($y=1$), while the remaining terms are labeled as 0($y=0$). Thus, we regard this auxiliary task as the binary classification task.

For generating the binary classification probability of each term, the Projector2 module is designed to combine a term's spatial context representations $H$ into a term binary classification probability, i,e., 
\begin{equation}
P_{term_i} = Sigmoid(w_2h_i + b_2)
\end{equation}
where $w_2$ and $b_2$ are linear weights and bias of Projector2 module, respectively. $h_i$ is $ith$ term spatial context representations from BERT.

\begin{table}
\centering
  \caption{The results of passage retrieval on MS Marco Dev. 'coCon': coCondenser, a technique that continually pre-trains BERT in an unsupervised manner. ‘Reranker taught’: distillation from a reranker.}
  \label{tab:sota}
  \begin{tabular}{ccccl}
    \toprule
    Method& Pre-trained model & Reranker taught & Hard negs & MRR@10\\
    \midrule
    \midrule
    Dense-vector Retriever & & &  & \\
    \midrule
    ANCE~\cite{xiong2020approximate}              & $RoBERTa_{base}$   &            &            & 33.8 \\
    ADORE~\cite{zhan2021optimizing}              & $RoBERTa_{base}$   &            & \checkmark & 34.7 \\
    TAS-B~\cite{hofstatter2021efficiently}              & $DistilBERT$       & \checkmark &            & 34.7 \\
    TCT~\cite{lin2021batch}                & $BERTbase$         & \checkmark &            & 33.5 \\
    TCT-ColBERT~\cite{lin2021batch}        & $BERTbase$         & \checkmark & \checkmark & 35.9 \\
    Condenser~\cite{gao2021condenser}          & $Condenser_{base}$ &            & \checkmark & 36.6 \\
    coCondenser~\cite{gao2021unsupervised}        & $coCon_{base}$     &            & \checkmark & 38.2 \\
    ColBERTv1~\cite{khattab2020colbert}          & $BERTbase$         &            &            & 36.0 \\
    ColBERTv2~\cite{santhanam2021colbertv2}          & $BERTbase$         & \checkmark & \checkmark & 39.7 \\
    PAIR~\cite{ren2021pair}               & $ERNIE_{base}$     & \checkmark &            & 37.9 \\
    RocketQA~\cite{qu2020rocketqa}           & $ERNIE_{base}$     & \checkmark &            & 37.0 \\
    RocketQAv2~\cite{ren2021rocketqav2}         & $ERNIE_{base}$     & \checkmark & \checkmark & 38.8 \\
    AR2~\cite{zhang2021adversarial}                & $coCon_{base}$     & \checkmark & \checkmark & 39.5 \\
    LED~\cite{zhang2023led}                & $BERTbase$     &             & \checkmark & \underline{40.2} \\
    COILcr~\cite{fan2023coilcr}            & $coCon_{base}$   &             & \checkmark & 37.0 \\
    \midrule
    \midrule
    Hybrid Retriever & & &  & \\
    \midrule
    CLEAR~\cite{gao2021complement}              & $BERTbase$         &            &            & 33.8 \\
    COIL-full~\cite{gao2021coil}          & $BERTbase$         &            &            & 35.5 \\
    Unifieruni-retrieval  & $coCon_{base}$     &            &            & \underline{40.7} \\
    \midrule
    \midrule
    Lexicon-base Retriever & & &  & \\
    \midrule
    DeepCT~\cite{dai2019context}              & $BERTbase$         &            &            & 24.3 \\
    RepCONC~\cite{zhan2022learning}            & $RoBERTa_{base}$   &            & \checkmark & 34.0 \\
    SPLADE-max~\cite{formal2021splade}         & $DistilBERT$       &            &            & 34.0 \\
    SPLADE-doc~\cite{formal2021splade}          & $DistilBERT$       &            &            & 32.2 \\
    DistilSPLADE-max~\cite{formal2021splade}    & $DistilBERT$       & \checkmark &            & 36.8 \\
    uniCOIL~\cite{lin2021few}          & $BERTbase$         &            & \checkmark & 35.2 \\
    TILDE~\cite{zhuang2021tilde}       &$BERTbase$         &            &  \checkmark & 29.5 \\
    TILDEv2~\cite{zhuang2021fast}          &$BERTbase$         &            &            & 34.1 \\
    SelfDistil~\cite{formal2022distillation}         & $DistilBERT$       & \checkmark & \checkmark & 36.8 \\
    EnsembleDistill~\cite{formal2022distillation}     & $DistilBERT$       & \checkmark & \checkmark & 36.9 \\
    Co-SelfDistill~\cite{formal2022distillation}      & $coCon_{base}$     & \checkmark & \checkmark & 37.5 \\
    Co-EnsembleDistill~\cite{formal2022distillation}  & $coCon_{base}$     & \checkmark & \checkmark  & 38.0 \\
    FecTek  & $coCon_{base}$     &             & \checkmark  & \underline{38.2} \\
    FecTek-Distill-MinLM  & $coCon_{base}$     & \checkmark  & \checkmark  & \textbf{38.7} \\
    
  \bottomrule
\end{tabular}
\end{table}

\subsection{Loss}
The text-level knowledge is acquired by the ranking objective with a contrastive loss. Given a query $q$ and a set of $n$ passages $D = \{ d^+, d^-_1, d^-_2, ..., d^-_{n-1}\}$  the training loss is
\begin{equation}
\mathcal{L}_{text-level} = -log\frac{e^{s(q, d^+)}}{e^{s(q, d^+)}+\sum_{d^-_j \in D} e^{s(q, d^-_j)}}
\end{equation}
The term-level loss, represented as $\mathcal{L}_{term-level}$, injects this term-level knowledge into our FecTek system through an additional projector, as represented below.

\begin{equation}
\mathcal{L}_{term-level-q} = \frac{1}{Q} \sum_i^Q ylog(P_{term_i}) + (1-y)log(1-P_{term_i})
\end{equation}
where $Q$ is the length of the query and $P_{term_i}$ is the binary classification probability of $ith$ term in query.

\begin{equation}
\mathcal{L}_{term-level-p} = \frac{1}{P} \sum_i^P ylog(P_{term_i}) + (1-y)log(1-P_{term_i})
\end{equation}
where $P$ is the length of the passage and $P_{term_i}$ is the binary classification probability of $ith$ term in passage.

\begin{equation}
\mathcal{L}_{term-level} = \mathcal{L}_{term-level-q} + \mathcal{L}_{term-level-p}
\end{equation}

The total loss for our FecTek is
\begin{equation}
\mathcal{L} = \mathcal{L}_{text-level}  + \mathcal{L}_{term-level}
\end{equation}

\section{Experiments}

\subsection{Implementation Details}
\subsubsection{Datasets \& Metrics}
Our FecTek is trained and evaluated on the MS MARCO passage ranking dataset~\cite{nguyen2016ms} within the full ranking setting. The training set contains extensive 8.8M passages along with their corresponding 0.5M pairs of queries and relevant passages. Additionally, the development(dev) set includes 6980 queries and their relevant labels. Just like other works~\cite{formal2022distillation,lin2021few,zhuang2021tilde}, we employed doc2query–T5~\cite{nogueira2019doc2query} as an expansion model to generate queries for each passage. Moreover, taking inspiration from the work~\cite{formal2022distillation}, we considered the recently released dataset of msmarco-hard-negatives dataset\footnote{\url{https://huggingface.co/datasets/sentence-transformers/msmarco-hard-negatives.}} as our negative samples to acquire the text-level knowledge. The msmarco-hard-negatives dataset has been meticulously crafted by combining the top-50 hard negatives mined from BM25 and a set of 12 diverse dense retrievers. Following previous works, we report $MRR@10$ for MS MARCO dev.

\subsubsection{Experimental Setups}
We initialized our models with the fantastic coCondenser-marco~\cite{gao2021unsupervised} (unsupervised continual pre-training from BERT-base~\cite{devlin2018bert}). We trained all models for 5 epochs using the AdamW optimizer. The learning rate was set to 2e-5, with a warm-up ratio of 0.1 and linear learning rate decay. For each query, we incorporated 1 positive example and 15 hard negatives. Each batch involved 8 queries implemented on MS MARCO passage dataset. Queries and passages were truncated to the first 64 and 192 tokens, respectively. We use 8 NVIDIA-V100 GPUs, each with 32GB of memory per GPU to train our model. Inspired by work~\cite{lin2021few}, after encoding the corpus, we effectively quantized the weights into 8 bits, yielding impactful scores. Similarly, we employed a similar quantization technique for query weights. With these adjustments in place, our FecTek seamlessly integrates with inverted indexes.
\subsubsection{Distillation Setups} We keep the parameters consistent with the above parameters. We experiment with two rerankers as the knowledge of distillation: MiniLM-L-12-v2\footnote{\url{https://huggingface.co/cross-encoder/ms-marco-MiniLM-L-12-v2.}} and beg-reranker\footnote{\url{https://huggingface.co/BAAI/bge-reranker-large.}}. It's important to note that two reranker models are not further trained.

\subsection{Comparison with State-of-the-arts}

As shown in Tab.~\ref{tab:sota}, our FecTek achieves new state-of-the-art performance metrics on MS Marco dev. In the absence of distillations from rerankers, our Lexicon-base retrieval surpasses previous methods, elevating the performance of the best sparse method in MRR@10 by an impressive margin (+3.0\%). Our FecTek outperforms several prominent dense-vector retrievers, such as ColBERTv1~\cite{khattab2020colbert}(+2.2\%), Condenser~\cite{gao2021condenser}(+1.6\%), RocketQA~\cite{qu2020rocketqa}(+1.2\%) and COILcr~\cite{fan2023coilcr}(+1.2\%). As added the distillation from reranker, our FecTek can achieve 38.7\% MRR@10, surpassing previous methods with distillations from rerankers by a notable margin(+0.7\%). 

\subsection{Ablation Study}
\subsubsection{Effectiveness of each Component} The contributions of different components of FecTek are listed in Tab.~\ref{tab:Effectiveness_each_component}. By introducing the FCM module, the MRR@10 performance is improved from 37.1 to 37.6. The TKGM module increases the MRR@10 from 37.1 to 37.9. Finally, by using both, the performance can be significantly improved from 37.1 to 38.2 for MRR@10 on MS Marco dev. The ablation studies in Tab.~\ref{tab:Effectiveness_each_component} verify the effectiveness of each module in our FecTek.
\begin{table}
\centering
  \caption{Ablation study on the effectiveness of each component on MS Marco dev.}
  \label{tab:Effectiveness_each_component}
  \begin{tabular}{ccccl}
  Component & FCM & TKGM & MRR@10 \\
  \toprule
   &            &            &  37.1 \\
   & \checkmark &            &  37.6 \\
   &            & \checkmark &  37.9 \\
   & \checkmark & \checkmark &  38.2 \\
  \bottomrule
\end{tabular}
\end{table}

\subsubsection{Effectiveness of Different Distillation Approaches} The effects of different distillation approaches for FecTek are summarized in Tab.~\ref{tab:Effectiveness_each_distll}. When the MiniLM-L-12-v2 reranker was employed for distillation,  the MRR@10 improved from 38.2 to 38.7. Similarly, when the beg-reranker was used for distillation, the MRR@10 improved from 38.2 to 39.2. Ablation studies presented in Tab.~\ref{tab:Effectiveness_each_distll} support the notion that utilizing a more powerful model as a reranker for distillation yields more pronounced performance gains.
\begin{table}
\centering
  \caption{Ablation study on the effectiveness of distillation on MS Marco dev.}
  \label{tab:Effectiveness_each_distll}
  \begin{tabular}{cccl}
  Method & Reranker model &  MRR@10 \\
  \toprule
   FecTek &    -        &  38.2 \\
   FecTek-Distill-MinLM      & MiniLM-L-12-v2 &  38.7 \\
   FecTek-Distill-beg        & beg-reranker            &  39.2 \\
  \bottomrule
\end{tabular}
\end{table}

\section{Conclusion}
In this paper, we have proposed an innovative approach named FecTek, which incorporates two key modules: the Feature Context Module (FCM) and the Term-level Knowledge Guidance Module (TKGM). The FCM enriches the contextual representation of term weights by leveraging BERT's embedding to assign dynamic weights to each element within the embedding. Moreover, our TKGM intelligently leverages the term-level knowledge guidance module (TKGM) to guide the modelling process of term weights. Our FecTek method outperforms previous approaches, taking the best Lexicon-base method to new heights with a remarkable improvement in MRR@10 (+3.0\%). Additionally, our FecTek surpasses some mainstream dense-vector retrievers with an improvement of more than 1.2\%. Furthermore, when integrated with the distillation from reranker, our FecTek achieves an impressive 38.7\% MRR@10.
The experimental results on the MS Marco benchmark undeniably demonstrate the superiority of our proposed method over previous state-of-the-art approaches.
\bibliographystyle{ieeetr}

\bibliography{references}

\end{document}